\documentclass{article}

\usepackage[preprint]{neurips_2025}

\usepackage[utf8]{inputenc} 
\usepackage[T1]{fontenc}    
\usepackage{hyperref}       
\usepackage{url}            
\usepackage{booktabs}       
\usepackage{amsfonts}       
\usepackage{nicefrac}       
\usepackage{microtype}      
\usepackage{xcolor}         

\usepackage{amssymb}
\usepackage{amsmath}
\usepackage{graphicx} 
\usepackage[skip=10pt plus1pt, indent=0pt]{parskip}
\usepackage{dsfont}
\usepackage{algorithm,algorithmic}
\usepackage{subcaption}
\usepackage[normalem]{ulem}
\usepackage{xurl}
\usepackage{booktabs}
\usepackage{multirow}
\usepackage{array}
\usepackage{placeins}

\newtheorem{mypro}{Problem}

\def \E {\mathbf{E}}

\def \G {\mathbf{G}}
\def \V {\mathbf{V}}

\def \X {\mathbf{X}}

\def \Z {\mathbf{Z}}

\def \h {\mathcal{h}}

\def \N {\mathcal{N}}

\def \H {\mathbf{H}}
\def \h {\mathbf{h}}
\def \W {\mathbf{W}}

\title{Robust Learning on Noisy Graphs via Latent Space Constraints with External Knowledge}

\author{%
  Chunhui Gu \\
  Department of Biostatistics \\
  University of Texas MD Anderson Cancer Center \\
  \texttt{chunhui.gu@mdanderson.org} 
  \And
  Mohammad Sadegh Nasr \\
  Department of Computer Science and Engineering \\
  University of Texas at Arlington \\
  \texttt{mohammadsadegh.nasr@mavs.uta.edu} 
  \And
  James P. Long \\
  Department of Biostatistics \\
  University of Texas MD Anderson Cancer Center \\
  \texttt{JPLong@mdanderson.org} 
  \And
  Kim-Anh Do \\
  Department of Biostatistics \\
  University of Texas MD Anderson Cancer Center \\
  \texttt{kimdo@mdanderson.org} 
  \And
  Ehsan Irajizad \\
  Department of Biostatistics \\
  University of Texas MD Anderson Cancer Center \\
  \texttt{EIrajizad@mdanderson.org} 
}

\begin{document}

\maketitle

\begin{abstract}
Graph Neural Networks (GNNs) often struggle with noisy edges. We propose Latent Space Constrained Graph Neural Networks (LSC-GNN) to incorporate external “clean” links and guide embeddings of a noisy target graph. We train two encoders—one on the full graph (target plus external edges) and another on a regularization graph excluding the target’s potentially noisy links—then penalize discrepancies between their latent representations. This constraint steers the model away from overfitting spurious edges. Experiments on benchmark datasets show LSC-GNN outperforms standard and noise-resilient GNNs in graph subjected to moderate noise. We extend LSC-GNN to heterogeneous graphs and validate it on a small protein–metabolite network, where metabolite–protein interactions reduce noise in protein co-occurrence data. Our results highlight LSC-GNN’s potential to boost predictive performance and interpretability in settings with noisy relational structures.

Code repository: https://anonymous.4open.science/r/latent-space-constraint-GNN-23B6/README.md
\end{abstract}

\section{Introduction}

Graph Neural Networks (GNNs) have demonstrated significant improvement in modeling graph-structured data from many disciplines \cite{Zitnik_2018, fan2019graphneuralnetworkssocial, kipf2018neuralrelationalinferenceinteracting}. A key distinction between GNNs and other deep learning frameworks is their ability to aggregate information using known node connections through a message-passing mechanism \cite{RN2153}. Message passing is powerful because it systematically aggregates and transforms information from each node’s neighbors, allowing the learned representations to capture both local features and broader structural context within the graph.

The performance of GNNs is highly dependent on the quality of provided edges (a.k.a. links) \cite{RN2213, dai2022robustgraphneuralnetworks}. Ideally, all these links should capture meaningful relationships between nodes. However, in real-world graphs, this is rarely the case. This could be a result of two major reasons. First, the misclassification between negative (non-existent) and positive links. In most moderate to large graphs, the number of possible links (in magnitude of $N^2$, $N$ is the number of nodes) is much higher than the number of actual links \cite{RN2214, li2023evaluatinggraphneuralnetworks}, leading to the pervasive false-positive links both in the originally provided links and the succeeding new link prediction task. Second, the existence of boundary links that reflect the authentic but highly specialized or uncommon relationships, making them outliers that are difficult to generalize from and potentially detrimental to the overall modeling task \cite{RN2213, baranwal2022graphconvolutionsemisupervisedclassification}. We used the term "noisy links" to refer to both cases, as distinguishing between them is often impractical in real-world scenarios.

In many cases, a target graph is naturally embedded as a subgraph of broader more comprehensive graph. For example, in social network, an organisation’s internal friendship graph forms just one slice of the global social graph. Likewise, in biology, one omics layer constitutes its own relational network and becomes part of a larger multi-layer biological network when combined with cross-omics links \cite{Zitnik_2018, menche2015uncovering}. While the relationships within the target graph may be unclear, their links with nodes under the larger graph are more studied and can be used to improve the prediction for the target graph. However, the optimal way to incorporate such additional information is unclear.

We aim to address the challenge of "noisy links" in graph data by investigating how such issues can be alleviated by embedding the target graph as a subgraph within a larger, more accurate, and well-defined graph. To achieve this, we propose a novel framework that constrains the latent space by incorporating additional links introduced by augmenting the target graph. This approach helps to improve model robustness against noisy links and guides the learned embeddings toward more meaningful representations especially in many biology setting where we often have high-potential but noisy relationships and validated data sources.

Our contributions are threefold: (1) we introduce an innovative technique for training models by mitigating the effects of noisy direct links through the incorporation of additional clean links as a form of regularization (LSC-GNN); (2) we extend the proposed approach to accommodate heterogeneous graph learning scenarios; and (3) we validate the effectiveness of our model using real-world protein-metabolite heterogeneous network data.

\section{Related works}

Various approaches have been proposed to mitigate the impact of noisy links in GNNs. One common strategy involves pre-processing the adjacency matrix by either down-weighting or removing noisy links based on node feature similarity \cite{xu2022ned, wu2019adversarialexamplesgraphdata, deng2022garnet} or by applying low-rank decompositions \cite{entezari2020all} to get an adjusted adjacency matrix as input. Some methods choose to assign different learnable weights (or adding/removing edges) based on hidden layer similarity \cite{pmlr-v139-liu21k, zhang2020gnnguard} or optimize graph structure in an end-to-end fashion for the final task \cite{wu2019adversarial, wang2019graphdefense, xu2019topology}.

Another strategy is to incorporate regularization terms directly into the training objective to enhance robustness in an end-to-end manner. For instance, Jin et al. (2020) \cite{jin2020graphstructurelearningrobust} introduce a regularization framework that refines graph structure by assessing edge reliability based on node feature similarity, while simultaneously enforcing low-rank and sparsity constraints on the learned adjacency matrix. This dual regularization strategy aims to refine the graph by preserving meaningful connections while filtering out noisy edges. Tang et al. (2020) \cite{Tang_2020} adopt a self-attention mechanism that enables the model to assign different attention coefficients to normal edges and perturbed ones. Their approaches utilize external clean graphs similar to the target poised graph with known perturbation and employ a regularization loss to widen the margin between mean values of two distributions.

Additional studies exploit the distribution of learned features in hidden layers, such as assigning lower weights to edges with higher variation \cite{elinas2020variational} or using a soft median aggregation that takes the distribution of learned feature into consideration \cite{geisler2021robustness}, thereby reducing the influence from outlier neighbors introduced by perturbed edges.

Our proposed LSC-GNN method differs from the above studies: i) Our approach regularizes using new external relationships, while the approaches above regularize based on information in the target graph. ii) While Tang et al. (2020) and our model both incorporate additional clean links to improve performance on the noisy graph, the way in which external information is applied differs significantly. iii) Existing studies primarily focus on homogeneous graph setting, whereas we demonstrate that our method extends naturally to heterogeneous graphs, which are especially common in the biology domain.
\section{Preliminaries}
\subsection{Notations}

A given graph $\G = (\V, \E, \X)$ is defined by its nodes (vertices). $\V = \{v_1, \dots, v_n\}$, set of edges (links) $\E = \{(v_i, v_j)\mid \text{if there is edge from } v_i \text{ to } v_j\}$, and $\X \in N \times F$ is the node feature matrix with each row corresponding to a node and each column corresponding to a feature ($F$ is the number of node features). For node $i$, we use $z_i$ to denote the latent space of dimension $L$ with $L < F$ and a $N \times L$ latent matrix $\Z$ for all nodes.

\subsection{Basic GNN model design}
In this subsection, we introduce the general architecture for learning node representations in a lower-dimensional latent space using a Graph Neural Network (GNN). We aim to map graph-structured data into a latent space that preserves essential topological and semantic relationships between nodes.

A graph encoder $f$ is a function that transforms input graph and node features into meaningful node embeddings. Formally, we denote the encoder by $f: \mathds{R}^F \rightarrow \mathds{R}^L$, where $F$ is the input feature dimension and $L$ is the dimension of learned latent space. This encoder can be instantiated by any message-passing GNN model, such as GAT \cite{velickovic2018graphattentionnetworks} and GIN \cite{xu2019powerfulgraphneuralnetworks}.

Most GNNs follow a message-passing framework in which each node aggregates information from its neighbors layer by layer, and each layer only considers first-order neighbor information. 

Suppose we have an K-layer GNN. Formally, the k-th GNN layer is defined as:

\begin{equation} 
\label{eq:general_layer} 
\h_i^{(k)} = \text{UPDATE}^k\Big(\h_i^{(k-1)}, \text{AGG}^k\big({\h_j^{(k-1)} \mid j \in \mathcal{N}_i}\big)\Big),
\end{equation}

Where $h_i^{(k)}$ is the learned node representation (hidden state) of node $i$ at the k-th layer and $\mathcal{N}_i$ is the first-order neighbor of $i$. $\text{AGG}^k(\cdot)$ is an learnable aggregation function that combines neighbor features and $\text{UPDATE}^k(\cdot)$ is another function that take the node representation based on its current state and the aggregated neighbor information at the $k$-the layer. 

By stacking multiple such layers, each node’s representation progressively captures higher-order relational information from its neighbors. After the final (i.e. K-th) layer, we obtain $\H^{(K)} = \{\h_1^{(K)}, \dots, \h_N^{(K)}\}$, which we denote as matrix $\Z$, which is the learned latent space of all nodes.
 
The learned latent space of nodes can be used for downstream tasks through a task-specific head. For example, in Kipf(2013), a simple inner-product decoder was used to reconstruct adjacency matrix $\boldsymbol{\hat{A}} = \sigma(\boldsymbol{Z \cdot Z^T})$, where $\sigma(\cdot)$ is the logistic sigmoid function. Besides, a multi-layer perceptron (MLP) can be used as a classifier take the latent space as input and output node label classification.

\subsection{Problem Definition}
The problem of leveraging external links introduced by considering the target graph as sub-graph under a large full graph for learning a robust GNN against noisy target graph is formally defined as:

\begin{mypro}
\textit{Given the \textbf{target graph} $\G_t =  (\V_t, \E_t, \X_t)$, where $\V_t$ is the set of nodes of interest and $\E_t$ is the set of edge with noise. Suppose there exists a larger graph $\G_f = (\V_f, \E_f, \X_f)$, such that $\G_t$ is a subgraph of $\G_f$ (i.e. $\V_t \subset\V_f$ and $\E_t \subset \E_f$. We denote the additional clean external edges as $\E_r = \E_f \setminus \E_t$ (in $\E_f$ not in $\E_t$) and any external nodes as $V_r = \V_f \setminus \V_t$}. 

\textit{Define the \textbf{regularization graph} $\G_r = (\V_r, \E_r, X_f)$, which isolates only the external edges and their associated nodes. Our objective is to learn a robust GNN that maps each node $v \in \V_t$ to latent representation $\boldsymbol{z_v} \in \mathds{R}^d$, such that the resulting latent space effectively mitigates the impact of noisy edges. In particular, we aim to leverage $\G_r$ to \textbf{constrain} the latent representations derived from $\G_f$, thereby reducing noise-related artifacts. The learned embeddings should support downstream tasks such as node classification and link prediction within the target graph $\G_t$.}
\end{mypro}
\section{Proposed framework}

\subsection{Constrained Latent Space Learning}
\subsubsection{External Links and Model Assumption} 
External links in a graph refer to relationships that are not present in the target graph but are introduced when new nodes are introduced. These links represent interactions between the newly introduced nodes and the existing nodes in the target graph.

\begin{figure}
    \centering
    \includegraphics[width=0.8\linewidth]{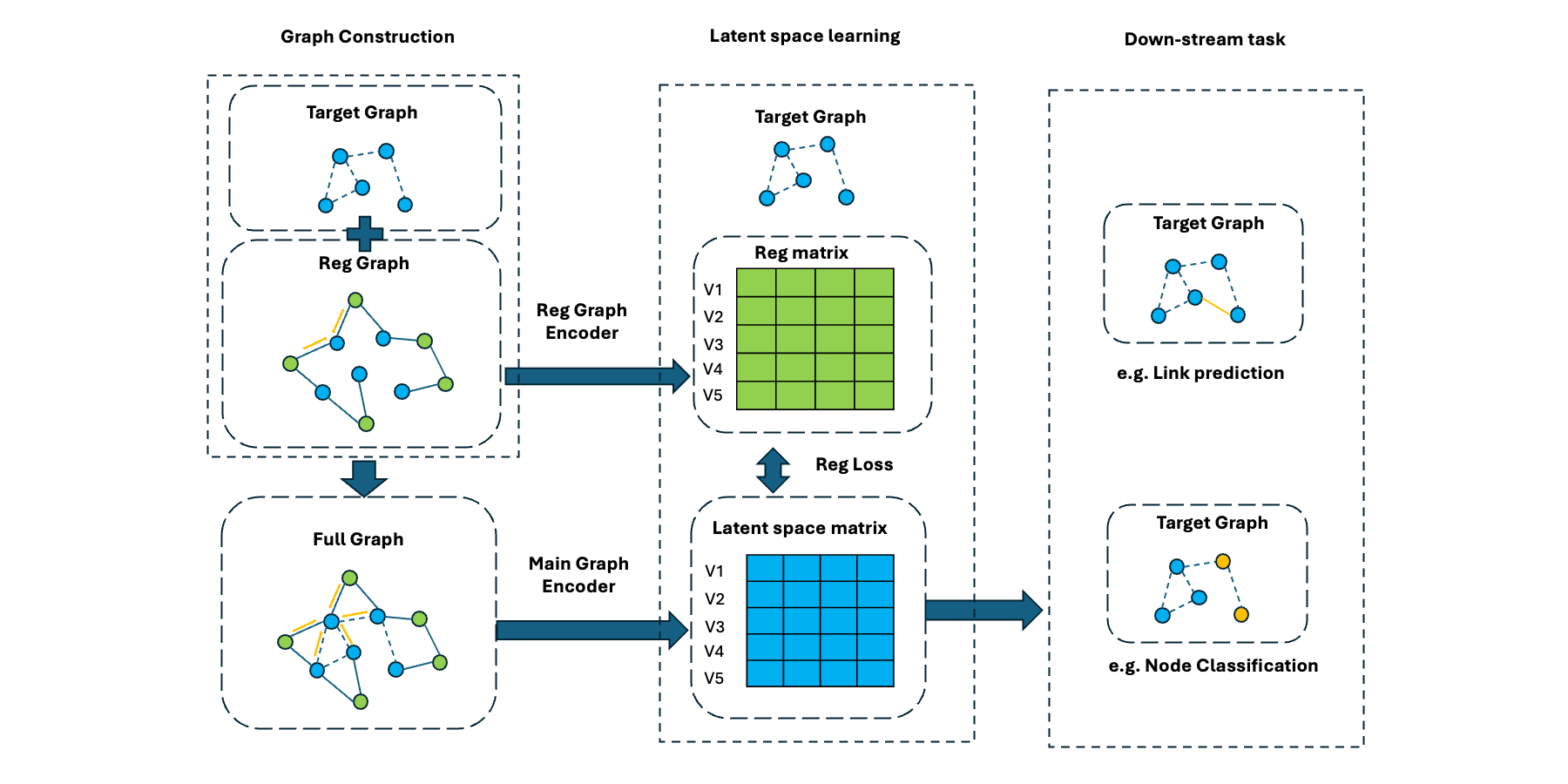}
    \caption{Framework of LSC-GNN: (1) The target graph contains noisy links (dashed lines), while the regularization graph is constructed using external nodes and clean edges (solid lines). The full graph is formed by combining both. (2) The main graph encoder learns a latent space matrix for nodes in the target graph using the full graph, while the regularization encoder learns a separate matrix using the regularization graph. The difference between these matrices is computed as the regularization loss. (3) The latent space matrix is processed by a task-specific head to obtain the target loss. During training, gradient descent minimizes a weighted sum of the regularization and target losses, while only the target loss is used for validation and testing.}
    \label{fig:framework}
\end{figure}

One natural approach to use the external links is to implement the full graph to train the model with loss traced only for tasks of the target graph. In our framework, however, external links are processed additional through a dedicated GNN encoder. These links can be used for regularization with different strengths. The model assumes that true meaningful links are more likely to form a densely connected cluster when embedded within a clean, larger graph than noisy links (Figure~\ref{fig:framework}). 

\subsubsection{Proposed Model}
\textbf{Three model components:} 
The default model has two component: a graph encoder $f: X \mapsto Z$ and a task-specific head $g: Z \mapsto Y$. The encoder $f$ operates on the full graph $\G_f$ to get latent space for nodes in the target graph $\G_t$. We now introduce a third component: a latent-space regularization encoder $f^\prime: X \mapsto Z^\prime$ to calculate the latent space matrix $\Z^\prime$ for nodes in the target graph using the regularization graph $\boldsymbol{G_r}$.

Specifically, we employ an encoder architecture based on the Graph Attention Network (GAT) \cite{velickovic2018graphattentionnetworks} due to its self-attention mechanism. This mechanism allows the model to dynamically adjust the importance of connections, assigning higher weights to edges supported by external connections provided via latent-space regularization.

The self-attention mechanism computes attention coefficient $e_{ij}$, which measure the importance of node $j$ to node $i$ during message passing, using:

\begin{equation}
\small\label{eqn:attn_coef}
    \text{e}_{ij}^k =
    \text{LeakyReLU} \big ( (\mathbf{a}^k)^\top [\W^k \h_i^k \| \W^k \h_j^k] \big ),
\end{equation}

where $\mathbf{a}^k$ and $\W^k$ are trainable parameters, and $\|$ represents the concatenation of vectors. The attention coefficients are restricted to first-order neighbors only. For a node $v_i$, $\text{a}_{ij}^k$ is computed based on $j \in \N_i$. These attention coefficients for $v_i$ are then normalized over all nodes $j \in \N_i$, ensuing comparability across different nodes:

\begin{equation}
\small
\label{eqn:attn_score}
    \alpha_{ij}^k = \text{softmax}_j(e_{ij}) = \frac{
    \text{exp}\big (\text{e}_{ij}^k \big )
    }{\sum_{k\in\N_i}
    \text{exp}\big (\text{e}_{ik}^k \big )
    }.
\end{equation}

The normalized attention coefficient combined with weight matrix generate a linear combination of node features. This aggregation process updates the node representations, forming the basis of message passing in the graph encoder. Specifically, a message-passing layer is defined as follows

\begin{equation} 
\small\label{eqn:gat}
    \h_i^{k} = \sigma \big ( \sum_{j\in \N_j} \alpha_{ij}^{k-1}  \W^{k-1} \h_j^{k-1} \big ).
\end{equation}

And latent space is defined as $\mathbf{z}_i = \h^{K}_i$, which is the output of the last message-passing layer. The loss function consists of two parts: the target loss and the regularization term.

\begin{equation}
    \mathcal{L}_{total} = \mathcal{L}_{target} + \lambda \cdot \mathcal{L}_{reg}
\end{equation}

The target loss is a binary cross-entropy loss $\mathcal{L}_{target}$ on node classification in the target graph $\boldsymbol{G_t}$ using learned latent space.

The regularization loss term is the mean square distance between latent spaces obtained by the GNN encoder using only the target graph and the GNN encoder using all links except those links within the target graph:

\begin{equation}
\mathcal{L}_{reg} = {\frac{1}{|V^t|L}} \sum_{ k \in V^t} \sum_{m = 1}^{L} (z_{k, m} - z^r_{k, m}) ^ 2
\end{equation}

where $\V^t$ is the nodes in the target graph $\boldsymbol{G_t}$, and $z_{k, m}$ is the $m$-th element of the latent space for node $k$. The regularization loss is normalized by the size of the final latent space $L$ and the number of nodes in the target graph $|V_t|$. And $\boldsymbol{Z^\prime}$ is the regularization latent space matrix output from $f'(\cdot)$:
\begin{equation}
    \boldsymbol{Z^\prime} = f'_{\boldsymbol{G_r}}(\boldsymbol{X, A_r})
\end{equation}

The regularization term can be interpreted as a measure of disagreement between the two graph encoders using full graph and regularization graph respectively. It imposes a higher penalty when the node similarities, as represented in the latent space based on external links, deviate from the similarity learned from the full graph. In other words, when the graph encoder that operates on the full graph begins to overfit to noisy links, the regularization term will prevent overfitting by penalizing representation learned from links without support from external connection and guiding to learned latent space towards more functionally relevant representations that align with known interactions within the larger, more reliable graph.

\begin{algorithm}
\caption{Latent Space Constrained Graph Neural Network}
\begin{algorithmic}[1]
  \scriptsize
  \STATE Given a target set of nodes $\mathbf{v}$ with noisy edges $\mathbf{e_v}$ and another set of nodes $\mathbf{u}$ with clean edges $\mathbf{e_u}$ and $\mathbf{e_{vu}}$.
  \STATE Construct the full graph $\mathbf{G_f} = (\mathbf{v} \cup \mathbf{u}, \mathbf{e_v} \cup \mathbf{e_u} \cup \mathbf{e_{vu}})$ and the regularization graph $\mathbf{G_r}$ by removing $\mathbf{e_v}$.
  \WHILE{not converged}
      \STATE Generate embeddings matrix $\mathbf{Z_f}$ from $\mathbf{G_f}$ using encoder $f(.)$.
      \STATE Generate embeddings matrix $\mathbf{Z_r}$ from $\mathbf{G_r}$ using encoder $f'(.)$.
      \STATE Compute reconstruction loss $\mathcal{L}_{\text{recon}}$ by decoding $\mathbf{Z_f}$ for downstream tasks.
      \STATE Compute regularization loss $\mathcal{L}_{\text{reg}}$ by comparing $\mathbf{Z_f}$ and $\mathbf{Z_r}$.
      \STATE Optimize total loss: $\mathcal{L}_{\text{total}} = \mathcal{L}_{\text{recon}} + \lambda \mathcal{L}_{\text{reg}}$.
      \STATE Update model parameters using gradient descent.
  \ENDWHILE
\end{algorithmic}
\end{algorithm}

\subsubsection{Simulation of dataset}
\textbf{Graph decomposition}: 
To simulate the concept of a target graph embedded within a larger graph, we partition the original graph into two subgraphs: the target graph and the regularization graph. A set of random sampled nodes along with all edges between them to form the target graph $\boldsymbol{G_t = (X, A^t)}$. The regularization graph $\boldsymbol{G_r = (X, A^r)}$ consists of the remaining nodes, the edges among them, and the edges connecting these nodes to those in the target graph. The target graph represents the portion of the network where link prediction is performed, while the regularization graph provides additional context and structural information without being directly involved in the prediction task.

\textbf{Adding false positive links:} After we got the target graph $\boldsymbol{G_t}$, false positive links were added by using negative sampling with a predefined ratio. We assumed the original links in the graph were the true links. After false-positive links are added to the target graph. We combine the target graph and regularization graph to get the full graph $\boldsymbol{G_f}$. In real setting, we don't need to add false positive links to get the full graph since it is the larger graph when additional nodes are introduced. We only need to get the regularization graph based on a set of edges considered stable.

\section{Experiments on node classification}

\textbf{Scenarios: } We tested the constrained latent spacing learning using various public datasets, including Cora, CiteSeer, and PubMed \cite{sen:aimag08}, and different false negative link ratios and regularization strengths. We used two-layer full-graph encoders with a 32-dim hidden size and a 16-dim latent size and a single-layer regularization encoder.

\begin{table}[ht]
\centering
\begin{tabular}{ccccc}
\hline
Dataset & Nodes & Edges & Features & Classes \\
\hline
Cora    & 2,708 & 10,556 & 1,433   & 7 \\
CiteSeer & 4,230 & 9,104  & 3,703  & 6 \\
PubMed   & 19,717 & 88,648 & 500    & 3 \\
\hline
\end{tabular}
\caption{Datasets}
\label{tab:my_label}
\end{table}

\textbf{Random node split:} Nodes in the target graph were split into train/validation/test sets \\
with a 70/10/20 ratio. The validation set is used for hyperparameter tuning.

\textbf{Model training: } We trained each model for 1000 epochs using Adam optimizer \cite{kingma2017adammethodstochasticoptimization} with a learning rate of 0.005. For LSCGNN models, we have the regularization coefficient $\lambda$ as an additional hyperparameter, and the model with best validation performance was evaluated on the test dataset. Each hyperparameter configuration was evaluated under 10 different random seeds. We included two classic GNN models (GCN and GAT) and three noise-resilient models (ProGNN, GCNJaccard, and GCNSVD). All models were configured to have the same number of layers, dimensionality, and other training hyperparameter for fair comparison.

\subsection{Results}
Results are presented in Table 2. LSC-GNN outperforms other models in most scenarios, even when only 10\% of external links are used when under low to medium perturbation rates. In some cases, GCN-Jaccard achieves better performance, possibly due to the influence of extreme outliers. Integrating Jaccard pre-processing with LSC-GNN could further enhance the model's performance.

\section{Ablation analysis}
We conduct an ablation study to assess the necessity of the regularization encoder using the Cora dataset with target ratio of 0.7 as an example. Specifically, we trained two two-component GNN models using the same architecture as in LSCGNN: one model took only the target graph $\G_t$ as input, whereas the other use only the regularization graph $\G_r$ as input. We observe that LSCGNN constantly outperform boths of these single-graph variants, as well as model using the full graph. These findings indicate that incorporating information from the regularization graph $\G_r$ enables the model to better distinguish meaningful signals from noise in target graph $\G_t$, a benefit that cannot be achieved when using the entire graph in its entirety or focusing on a single graph alone.

\section{Model complexity analysis}
 The time complexity for the graph encoder taking full graph as input is $O(|V_f|F^2 + |E_f|F)$ \cite{velivckovic2017graph}, where F is the number of input feature. The time complexity for the regularization encoder taking the regularization graph is $O(|V_f|F^2 + |E_r|F)$. The overall cost is roughly doubled compared to a single‐encoder GNN, but remains in the same asymptotic class.

\begin{figure}[H]
    \centering
    \begin{minipage}{0.48\textwidth}
        \centering
        \includegraphics[width=\linewidth]{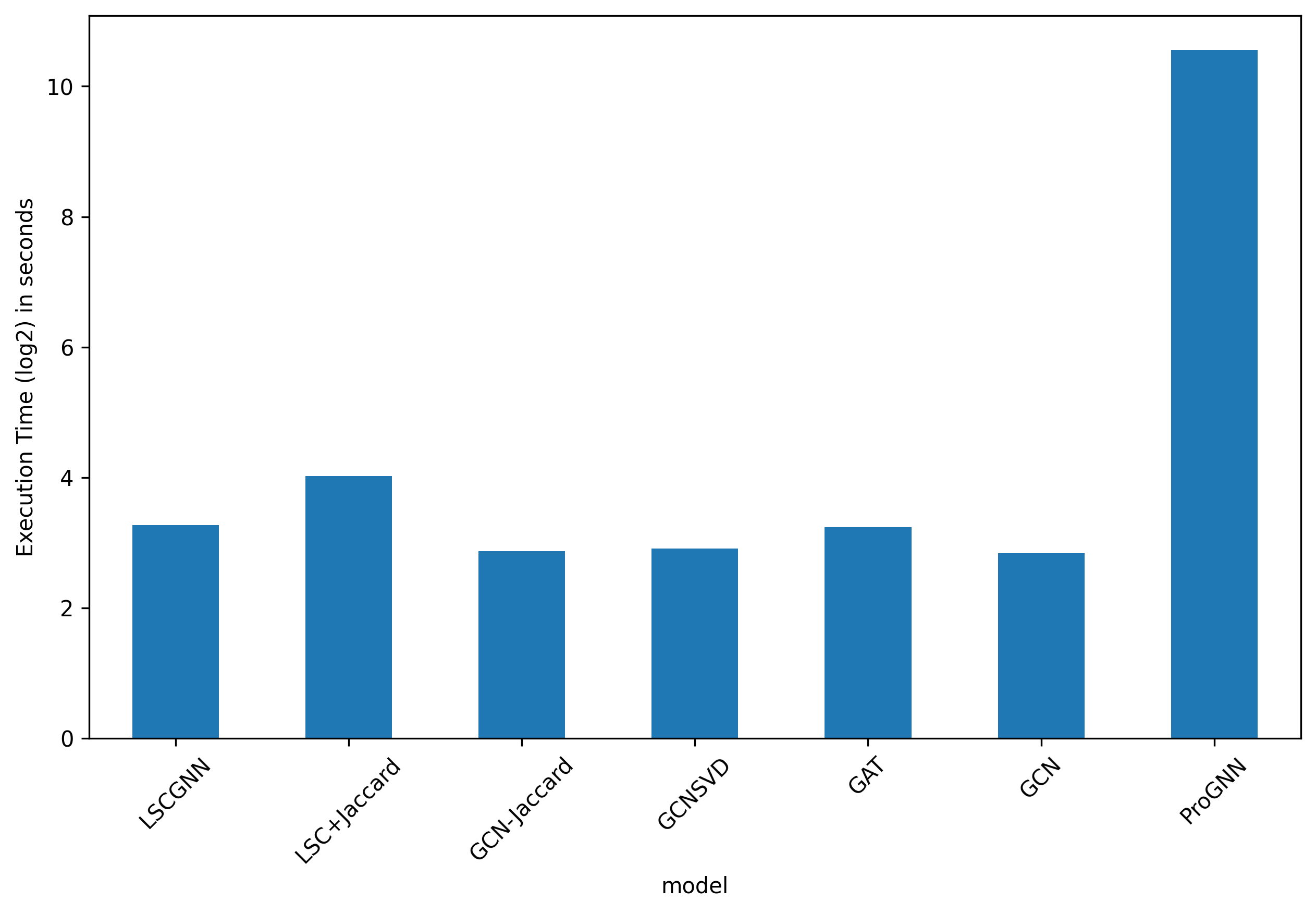}
        \caption{Runtime of different models. Trained on Nvidia V100-16G.}
        \label{fig:runtime}
    \end{minipage}
    \hfill
    \begin{minipage}{0.48\textwidth}
        \centering
        \includegraphics[width=\linewidth]{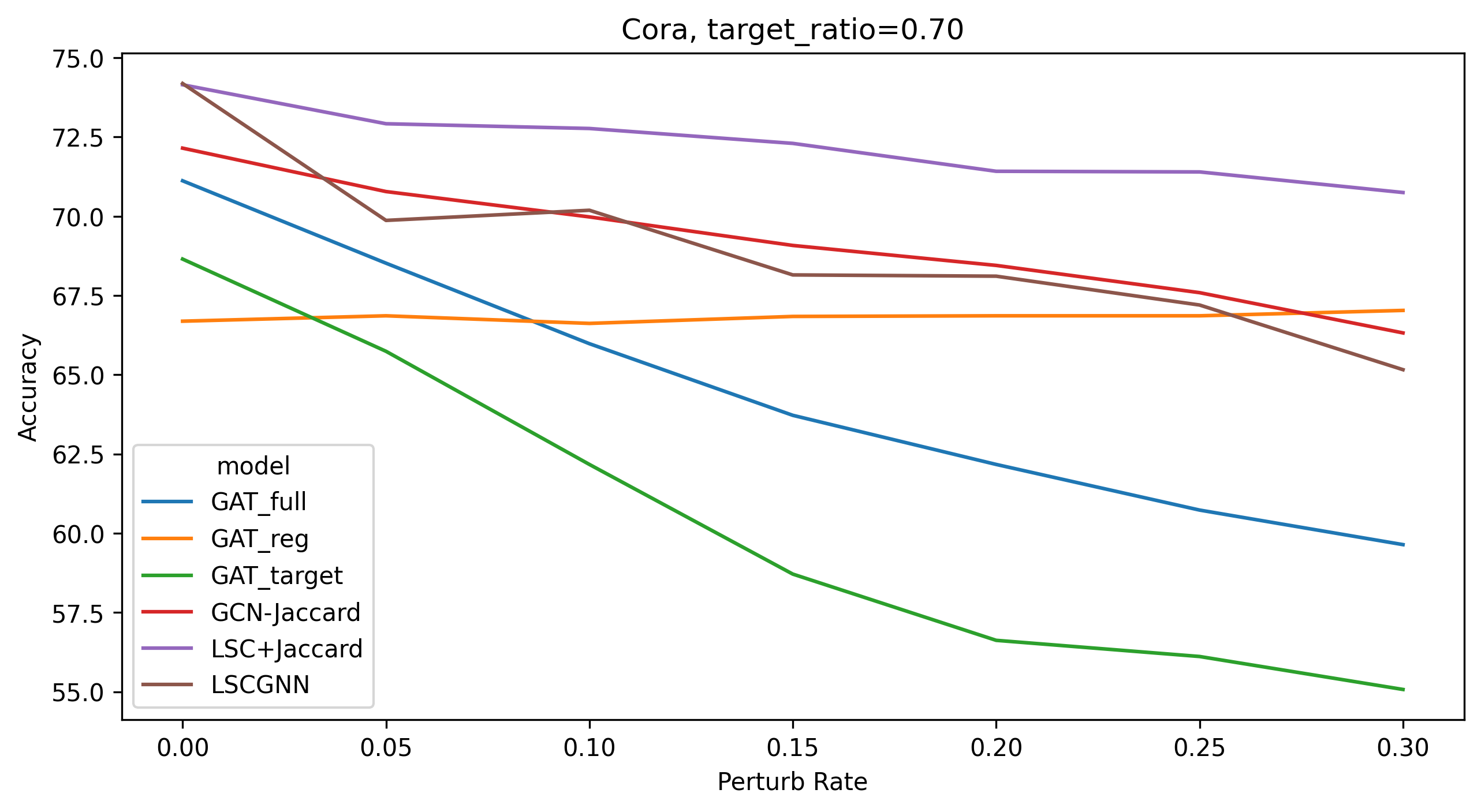}
        \caption{Ablation analysis.}
        \label{fig:ablation}
    \end{minipage}
    \label{fig:combined}
\end{figure}

\begin{table*}[ht]
\centering
\tiny
\caption{The model performance (ROC-AUC) under different settings using random perturbation. The columns indicate the rate of perturbation. ProGNN performance is unavailable on the PubMed dataset due to its cubic memory complexity, which makes it impractical for large graphs}
\begin{tabular}{llllllllll}
\toprule
 &  &  & 0.0 & 0.05 & 0.1 & 0.15 & 0.2 & 0.25 & 0.3 \\
Ratio & Dataset & Model &  &  &  &  &  &  &  \\
\midrule
\multirow[t]{20}{*}{0.50} & \multirow[t]{7}{*}{Cora} & GCN & 79.11±1.59 & 78.22±2.04 & 76.89±2.18 & 75.67±1.90 & 74.93±1.82 & 72.26±1.77 & 73.04±2.09 \\
 &  & GAT & 79.26±1.31 & 76.63±1.47 & 77.11±1.19 & 76.11±2.32 & 73.48±1.88 & 72.59±2.00 & 71.48±2.55 \\
 &  & GCN-Jaccard & 77.60±1.84 & 77.59±2.03 & 76.89±2.45 & 75.33±2.02 & 75.56±1.78 & 74.59±2.11 & 73.37±2.05 \\
 &  & GCNSVD & 77.74±0.36 & 76.04±0.45 & 76.15±0.57 & 75.93±0.61 & 73.81±0.59 & 71.78±0.81 & 70.04±0.65 \\
 &  & ProGNN & \textbf{81.52±0.32} & \textbf{82.44±0.70} & \textbf{80.44±0.38} & 79.00±0.70 & 76.85±1.21 & 76.19±1.01 & 73.48±1.11 \\
 &  & LSCGNN & 78.0±1.47 & 77.85±1.38 & 77.19±1.05 & \textbf{82.96±1.32} & \textbf{79.41±2.07} & 76.63±1.28 & \textbf{80.44±1.38} \\
 &  & LSC+Jaccard & 77.6±1.43 & 76.78±0.65 & 77.22±1.04 & 77.04±1.93 & 77.48±1.88 & \textbf{76.67±1.65} & 75.93±1.94 \\
\cline{2-10}
 & \multirow[t]{7}{*}{CiteSeer} & GCN & 69.04±0.83 & 67.83±1.03 & 65.51±1.64 & 64.79±1.63 & 62.14±1.39 & 61.93±0.73 & 61.66±1.12 \\
 &  & GAT & 68.34±1.48 & 67.47±1.46 & 64.46±1.25 & 63.46±2.29 & 60.66±1.62 & 59.79±1.87 & 61.24±2.08 \\
 &  & GCN-Jaccard & 69.88±1.05 & 69.58±1.08 & 67.08±1.20 & 67.08±1.54 & 66.06±1.24 & 65.21±1.40 & 64.40±1.44 \\
 &  & GCNSVD & 71.08±0.48 & 70.39±0.37 & 69.34±0.48 & \textbf{69.61±0.40} & 68.83±0.35 & 67.32±0.55 & 65.69±0.43 \\
 &  & ProGNN & \textbf{74.70±0.30} & \textbf{73.29±0.17} & \textbf{70.58±0.17} & 68.37±1.21 & 67.57±0.46 & 65.96±0.80 & 64.76±1.21 \\
 &  & LSCGNN & 70.03±1.07 & 71.2±1.5 & 70.09±1.03 & 69.19±1.21 & \textbf{69.97±1.16} & 65.21±2.08 & 64.31±1.34 \\
 &  & LSC+Jaccard & 70.54±0.63 & 69.76±1.32 & 68.98±0.59 & 68.97±1.4 & 68.44±1.32 & \textbf{68.92±1.53} & \textbf{68.4±1.22} \\
\cline{2-10}
 & \multirow[t]{6}{*}{PubMed} & GCN & 83.74±0.58 & 81.55±0.63 & 80.78±0.39 & 79.68±0.57 & 79.17±0.62 & 78.58±0.60 & 77.58±0.54 \\
 &  & GAT & 83.65±1.06 & 81.84±0.95 & 80.95±0.78 & 80.58±0.98 & 80.54±1.30 & 79.74±1.30 & 79.07±0.80 \\
 &  & GCN-Jaccard & 83.91±0.62 & 81.54±0.48 & 80.54±0.50 & 79.72±0.51 & 79.85±0.53 & 78.49±0.52 & 77.64±0.53 \\
 &  & GCNSVD & 85.26±0.12 & \textbf{84.97±0.18} & \textbf{84.38±0.14} & \textbf{84.04±0.15} & \textbf{83.54±0.19} & \textbf{83.38±0.13} & \textbf{82.97±0.11} \\
 &  & LSCGNN & 85.93±0.42 & 84.63±0.51 & 82.9±1.07 & 81.83±1.2 & 80.85±2.32 & 81.62±1.09 & 80.7±1.33 \\
 &  & LSC+Jaccard & \textbf{86.12±0.68} & 84.74±0.68 & 83.47±1.18 & 82.34±1.62 & 80.97±4.41 & 80.87±2.07 & 80.37±3.93 \\
\cline{1-10} \cline{2-10}
\multirow[t]{20}{*}{0.70} & \multirow[t]{7}{*}{Cora} & GCN & 82.64±0.59 & 81.14±1.28 & 80.16±1.12 & 79.34±1.20 & 77.52±1.44 & 74.62±1.44 & 73.22±1.05 \\
 &  & GAT & 82.72±1.21 & 80.34±1.47 & 79.05±1.83 & 77.92±2.19 & 75.17±2.17 & 73.01±1.46 & 70.63±2.10 \\
 &  & GCN-Jaccard & 81.53±0.83 & 81.06±0.76 & \textbf{80.50±0.38} & \textbf{79.74±0.94} & 78.63±1.36 & 76.99±1.46 & 76.10±1.38 \\
 &  & GCNSVD & 80.24±0.35 & 78.79±0.47 & 76.86±0.40 & 76.10±0.56 & 76.65±0.49 & 73.40±0.40 & 72.82±0.40 \\
 &  & ProGNN & \textbf{83.27±0.38} & 81.64±0.54 & 79.68±0.70 & 79.55±0.71 & 78.15±0.53 & 76.52±0.53 & 74.70±0.55 \\
 &  & LSCGNN & 81.03±2.19 & \textbf{82.72±1.24} & 79.87±1.66 & 79.21±2.18 & \textbf{81.24±2.0} & 76.41±1.35 & 75.33±1.13 \\
 &  & LSC+Jaccard & 79.89±1.5 & 79.26±1.71 & 78.21±2.23 & 78.73±2.32 & 78.15±1.95 & \textbf{78.31±2.05} & \textbf{78.02±1.8} \\
\cline{2-10}
 & \multirow[t]{7}{*}{CiteSeer} & GCN & 72.24±0.79 & 70.11±1.39 & 67.87±1.10 & 65.72±1.57 & 64.56±1.77 & 64.35±1.78 & 63.16±2.23 \\
 &  & GAT & 71.12±1.31 & 68.52±1.01 & 65.98±1.71 & 63.72±2.25 & 62.17±2.23 & 60.73±2.48 & 59.64±2.22 \\
 &  & GCN-Jaccard & 72.15±0.84 & 70.78±1.09 & 69.98±1.10 & 69.08±1.14 & 68.45±1.36 & 67.59±1.68 & 66.32±1.74 \\
 &  & GCNSVD & 71.01±0.42 & 69.20±0.32 & 67.59±0.30 & 66.71±0.34 & 65.33±0.34 & 64.62±0.30 & 65.08±0.33 \\
 &  & ProGNN & 70.39±0.13 & 69.03±0.22 & 66.17±0.33 & 65.59±0.57 & 64.95±0.22 & 63.44±0.38 & 62.58±0.43 \\
 &  & LSCGNN & \textbf{74.19±1.07} & 69.87±1.46 & 70.19±1.42 & 68.15±1.78 & 68.11±1.27 & 67.2±1.59 & 65.16±2.43 \\
 &  & LSC+Jaccard & 74.15±0.85 & \textbf{72.92±0.7} & \textbf{72.77±0.94} & \textbf{72.3±0.95} & \textbf{71.42±0.95} & \textbf{71.4±0.71} & \textbf{70.75±1.0} \\
\cline{2-10}
 & \multirow[t]{6}{*}{PubMed} & GCN & 84.33±0.50 & 82.04±0.72 & 80.57±0.77 & 79.99±0.69 & 78.29±0.73 & 78.34±1.31 & 78.31±1.00 \\
 &  & GAT & 83.45±0.77 & 81.56±0.77 & 80.49±0.82 & 80.13±0.84 & 78.95±0.66 & 78.35±0.71 & 78.04±0.86 \\
 &  & GCN-Jaccard & 84.23±0.41 & 82.05±0.77 & 80.55±0.80 & 80.09±0.56 & 78.58±0.55 & 78.53±0.96 & 78.38±1.12 \\
 &  & GCNSVD & 84.57±0.17 & 83.70±0.11 & 83.06±0.13 & 82.23±0.12 & \textbf{81.31±0.15} & 80.75±0.16 & 79.99±0.22 \\
 &  & LSCGNN & \textbf{86.0±0.6} & \textbf{85.37±0.69} & \textbf{83.54±2.11} & 81.92±1.35 & 81.21±1.21 & \textbf{81.11±0.86} & \textbf{81.09±1.01} \\
 &  & LSC+Jaccard & 85.3±1.4 & 83.15±3.96 & 82.1±3.16 & \textbf{82.68±1.16} & 81.11±1.69 & 80.68±1.67 & 80.06±1.63 \\
\cline{1-10} \cline{2-10}
\multirow[t]{20}{*}{0.90} & \multirow[t]{7}{*}{Cora} & GCN & 87.13±0.76 & 84.07±0.80 & 82.03±1.22 & 80.84±1.16 & 78.63±0.96 & 76.59±1.77 & 76.18±2.37 \\
 &  & GAT & 87.27±0.89 & 83.84±1.00 & 81.85±1.69 & 79.57±1.69 & 76.94±1.90 & 74.54±2.97 & 73.74±1.79 \\
 &  & GCN-Jaccard & 85.48±0.71 & 84.23±0.80 & \textbf{83.53±1.34} & 81.68±1.40 & 80.14±1.38 & 79.77±1.62 & 79.22±1.21 \\
 &  & GCNSVD & 81.21±0.17 & 79.20±0.46 & 77.90±0.34 & 75.63±0.48 & 72.84±0.47 & 71.23±0.63 & 69.39±0.38 \\
 &  & ProGNN & \textbf{88.40±0.24} & \textbf{85.61±0.40} & 81.29±1.09 & 78.01±0.73 & 74.33±0.53 & 71.36±1.28 & 68.34±1.08 \\
 &  & LSCGNN & 85.73±1.74 & 83.41±1.31 & 76.98±3.99 & 80.96±1.28 & 77.13±1.53 & 76.41±2.46 & 75.71±4.09 \\
 &  & LSC+Jaccard & 85.77±1.5 & 83.92±1.36 & 83.31±1.46 & \textbf{82.46±1.82} & \textbf{82.46±1.75} & \textbf{82.59±1.39} & \textbf{82.01±1.33} \\
\cline{2-10}
 & \multirow[t]{7}{*}{CiteSeer} & GCN & 72.12±0.65 & 68.43±0.94 & 68.15±0.95 & 67.11±1.44 & 65.08±1.57 & 63.47±1.95 & 62.05±1.89 \\
 &  & GAT & 72.02±0.87 & 68.16±0.52 & 66.98±1.28 & 65.46±1.65 & 63.20±1.08 & 60.47±1.61 & 58.93±1.91 \\
 &  & GCN-Jaccard & 72.52±0.65 & 70.08±0.52 & 69.32±0.84 & 68.16±0.78 & 66.23±0.99 & 65.79±1.22 & 64.71±0.93 \\
 &  & GCNSVD & 72.39±0.35 & 71.35±0.38 & 69.22±0.35 & 69.40±0.31 & 67.28±0.50 & 65.19±0.26 & 63.49±0.32 \\
 &  & ProGNN & 73.01±0.51 & 71.12±0.17 & 69.17±0.09 & 67.11±0.34 & 64.77±0.73 & 61.88±0.77 & 61.60±0.34 \\
 &  & LSCGNN & 73.84±1.05 & \textbf{73.72±1.35} & 70.07±1.04 & 71.6±1.35 & 68.6±1.46 & 68.2±1.71 & 65.84±1.56 \\
 &  & LSC+Jaccard & \textbf{73.92±1.22} & 73.22±0.95 & \textbf{72.64±1.27} & \textbf{72.07±1.11} & \textbf{71.44±1.4} & \textbf{70.89±1.3} & \textbf{70.6±1.64} \\
\cline{2-10}
 & \multirow[t]{6}{*}{PubMed} & GCN & 86.61±0.67 & 83.09±0.17 & 80.95±0.43 & 79.45±0.84 & 78.80±0.70 & 78.35±1.12 & 77.62±0.97 \\
 &  & GAT & 85.21±0.37 & 82.24±1.04 & 81.30±0.81 & 79.79±1.05 & 78.61±1.01 & 77.58±0.67 & 77.58±0.65 \\
 &  & GCN-Jaccard & 86.37±0.59 & 83.27±0.35 & 81.60±0.88 & 80.12±1.54 & 77.12±6.68 & 78.33±0.85 & 78.20±0.97 \\
 &  & GCNSVD & 86.78±0.13 & 85.37±0.17 & \textbf{84.69±0.15} & \textbf{83.53±0.21} & \textbf{82.60±0.14} & \textbf{81.69±0.20} & \textbf{80.70±0.15} \\
 &  & LSCGNN & \textbf{86.9±1.41} & \textbf{85.4±0.87} & 84.44±1.08 & 81.76±3.78 & 81.02±2.42 & 79.59±1.96 & 77.8±2.03 \\
 &  & LSC+Jaccard & 86.66±2.59 & 83.63±1.36 & 82.75±1.15 & 80.77±1.3 & 79.84±0.83 & 79.12±1.37 & 78.03±0.98 \\
\cline{1-10} \cline{2-10}
\bottomrule
\end{tabular}

\label{tab:results}
\end{table*}
\section{Expansion to heterogeneous graph}
Although the framework described thus far focuses on homogeneous graphs, where all nodes and edges share a single modality, many real-world applications—such as multi-omics biological networks, knowledge graphs, and recommendation systems—feature multiple node and edge types. The latent space regularization strategy introduced above can be naturally adapted to these heterogeneous settings.

\subsection{Challenges in Heterogeneous Graphs}
Heterogeneous graphs consist of multiple node and edge types. In a biological setting, for instance, a single network may include protein nodes, metabolite nodes, and various edges (e.g., protein–protein interactions or metabolite–protein relationships). Extending the constrained latent space approach to such graphs poses several challenges: 1) Each relationship type (e.g., “interacts with,” “catalyzes”) may imply distinct semantic or structural information. 2) Nodes may exhibit different feature sets or dimensionalities (e.g., amino acid sequence embeddings vs. gene expression vectors), complicating uniform processing and integration. 3) Certain node types can form dense subgraphs, while others remain sparsely connected, which increases the complexity of graph-based message passing and embedding aggregation.

\subsection{Adapting the Latent Space Constraint Framework}
To illustrate the application of our proposed method in heterogeneous graphs, we begin with a simplified case involving two node types, A and B. This setup naturally extends to more complex heterogeneous graphs involving multiple node and edge types. In this example, the graph contains four possible relationships: A-A, B-B, and cross-type edges (A-B and B-A).

\textbf{Target graph and regularization graph}: Consistent with the homogeneous case, we designate the graph consist of only node type B as the target graph. The regularization graph is then formed by retaining all nodes but omitting B-B edges, thereby preserving the relationships between among type A nodes and between A and B nodes (A-B, B-A). Under this formulation, the full graph includes all nodes and edges, the target graph is restricted to B-B edges, and the regularization graph includes the remaining edges that do not directly connect type B nodes.

\textbf{Encoder setup}: Hetergenous graph, however, introduce multiple node and edge types, requiring more flexible message-passing architecture. One straightforward strategy is to implement separate sub-encoders, each specialized for particular edges type and then aggregate the sub-encoders' output at the node level. Alternatively, models explicity designed for heterogenous networks, such as R-GNC and HGT can be used \cite{schlichtkrull2017modelingrelationaldatagraph} or HGT \cite{hu2020heterogeneousgraphtransformer}. The latent space constraint framework remains applicable regardless of the choice of backbone model. With a clear definition of target graph, regularization graph, full graph, and graph encoder in the heterogeneous setting, the same latent space constraint principles can be effectively implemented.

\section{Small Case Study: Protein-metabolite heterogeneous network}
Proteins are essential to cellular processes, and protein–protein interactions (PPIs) form the backbone of many biological mechanisms. PPIs vary under different physiological conditions, such as cellular localization and post-translational modifications, and high-throughput experiments often yield high false-positive rates due to artificial assay conditions. An alternative perspective, protein co-occurrence, captures functional or biological relationships even in the absence of direct physical binding. However, co-occurrence data can also be noisy, arising from biological variability, experimental artifacts, and computational biases.

In contrast, metabolite–protein interactions (MPIs) often undergo direct biochemical validation and are meticulously curated in databases such as KEGG \cite{kanehisa2000kegg}. Their reliability and strong experimental grounding make them a valuable anchor for guiding protein-related predictions. By integrating high-confidence MPI data, one can potentially reduce the impact of noise in large-scale PPI datasets and improve the overall biological interpretability of the resulting models.

\subsection{Dataset}
We constructed a heterogeneous graph by combining high-quality MPI data from KEGG and PPIs from STRING database. We used 1024-dimensional embeddings generated by ProtT5 prottrans\_t5\_xl\_u50 model \cite{brandes2022proteinbert}, a large language model for protein amnio-acid sequence (downloaded from UniProt)  as the feature vector. For metabolite, we used RDKit (https://www.rdkit.org/) to generate Morgan molecular fingerprints (a.k.a extended-connectivity fingerprint ECFP) based on the SMILES string of each molecule, which yielded a binary vector of 1024 dimensions \cite{landrum2013rdkit} and further reduced to 128 dimensions using principal component analysis (PCA). While there other algorithms available for generating protein and metabolite feature vector, the two we used are recommended or demonstrate better performance from previous studies \cite{bioembeddings, wang2023mpi}.

\subsection{Proposed Model}
We built two heterogeneous graph encoders using SAGEConv message-passing from GraphSAEG. Using PyTorch Geometric package \cite{FeyLenssen2019} using the idea of replicating message-passing layers for each unique edge type. Both encoders consist of two layers, with the hidden and latent dimensions set to 32 and 64, respectively. Multiple edge types are aggregated at the node level using a summation operator.

\subsection{Model training}
We trained for maximum 1000 epochs using Adam \cite{kingma2017adammethodstochasticoptimization} with a learning rate of 0.005. The dataset is split into 50\% train, 20\% validation, and 30\% test simulate a real-world scenario in which only approximately half of the relationships typically known in advance. The model with best validation performance in terms of ROC-AUC was evaluation on the test dataset.

\subsection{Results}
1. Using only protein-protein occurrence: ROC-AUC 0.92.
2. Using both protein-protein occurrence and MPI (full graph): ROC-AUC 0.94. 3. Using full graph with regularization: ROC-AUC 0.96.

\section{Conclusion}
We introduce Latent Space Constrained Graph Neural Networks (LSC-GNN) to address the challenge of noisy links in graph-based learning. Our framework leverages the structural information from more reliable external links to regularize the latent space of a noisy target graph, thereby improving the robustness and generalizability of models.

Extensive experiments on benchmark datasets demonstrate that the proposed approach effectively mitigates the impact of noisy links. By incorporating external connections as a regularization constraint, LSC-GNN achieves superior performance compared to classic GNN models and other noise-resilient methods in graph with moderate noisiness. Additionally, the framework naturally extends to heterogeneous graph settings, making it well-suited for complex heterogeneous networks characterized by multiple entity types and diverse relationships. As we have shown, this approach can improved accuracy in biological network modeling, which may lead to better understanding of diseases or drug targets.  

\section{Limitations} While this study demonstrates promising results through the incorporation of external nodes and links, in real-world scenarios, such high-quality auxiliary data may be unavailable, noisy, incomplete, non-informative with respect to the target graph, or only partially overlapping with it. Since the models are evaluated using incremental regularization strengths on the training set, and the best model is selected based solely on the final target loss in the validation set, the procedure avoids assigning a non-zero coefficient to the regularization term unless it is beneficial, thereby guaranteeing no detrimental effect on the final results.

\bibliography{cgvae}
\bibliographystyle{unsrt}
\appendix

\section{Technical Appendices and Supplementary Material}

\
We use IBM Load Sharing Facility (LSF) system in Linux to execute the experiment using V100-16G GPU with 10 CPU cores and 160G memory.

\textbf{Negative Societal Impacts}: LSC-GNN could be applied to infer associations between researchers, employees, or organizations within a confidential or proprietary network (e.g., inferring co-authorship or collaboration within a private R\&D consortium), potentially violating privacy or intellectual property.

\textbf{Funding}: 
This work was supported by National Institutes of Health [grant number U01CA271888 to S.H., RP160693 to K.A.D.]; Specialized Programs of Research Excellence (grant number P50CA140388 to K.A.D. and J.P.L.); the Center for Clinical and Translational Science (grant number TR000371 to K.A.D. and J.P.L.); and the generous philanthropic contributions to The University of Texas MD Anderson Cancer Center Moon Shots Program.

\textbf{Conflict of Interest}: None declared.

\newpage
\section*{NeurIPS Paper Checklist}
\begin{enumerate}

\item {\bf Claims}
    \item[] Question: Do the main claims made in the abstract and introduction accurately reflect the paper's contributions and scope?
    \item[] Answer: \answerYes{} 
    \item[] Justification: 
    The main claims in the abstract and introduction are well-aligned with the paper’s contributions and scope. The introduction clearly enumerates the key contributions of the LSC-GNN framework — specifically: (1) using external “clean” links to regularize the latent space and mitigate the effect of noisy edges, (2) extending this approach to handle heterogeneous graph settings, and (3) validating the method on a real-world protein–metabolite heterogeneous network. In the body of the paper, each of these claims is substantiated by the content. The method section introduces the dual-encoder architecture with a latent space regularization loss, demonstrating how external clean links are incorporated to guide node embeddings — directly supporting the claim that LSC-GNN leverages external knowledge to combat noisy edges. In the result section, we show that LSC-GNN outperforms standard GNN baselines and noise-resilient variants on benchmark datasets with a moderate perturbation. Furthermore, the paper extends the framework to heterogeneous graphs as claimed. A dedicated experiment on a protein–metabolite network (a heterogeneous graph scenario) is included, and the positive results there validate that the method works as intended beyond just homogeneous graphs. In sum, the contributions listed in the introduction (latent space regularization with external links, noise robustness, and heterogeneous graph support) are all clearly reflected in the subsequent sections of the paper.

\item {\bf Limitations}
    \item[] Question: Does the paper discuss the limitations of the work performed by the authors?
    \item[] Answer: \answerYes{} 
    \item[] Justification: We discuss the limitation of approach in the conclusion section. The approach is limited to the existence of a high-quality external graph that are complete, informative to the target graph, and share reasonable overlapping with the target graph. However, we also mention that due to the adjustable regularization strength, this approach is guaranteed to have non-detrimental effect. 

\item {\bf Theory assumptions and proofs}
    \item[] Question: For each theoretical result, does the paper provide the full set of assumptions and a complete (and correct) proof?
    \item[] Answer: \answerNA{} 
    \item[] Justification: \answerNA{}

    \item {\bf Experimental result reproducibility}
    \item[] Question: Does the paper fully disclose all the information needed to reproduce the main experimental results of the paper to the extent that it affects the main claims and/or conclusions of the paper (regardless of whether the code and data are provided or not)?
    \item[] Answer: \answerYes{} 
    \item[] Justification: We provides sufficient detail to reproduce the main experimental results and validate its key claims. We describes the data sources, graph construction procedures, model architecture (including the dual-encoder setup), loss functions, training procedures, hyperparameters, and evaluation metrics in details. For benchmark datasets, standard splits and perturbation protocols are used, and for the heterogeneous case study, we specifies the protein and metabolite embeddings, feature preprocessing, and model configuration.

\item {\bf Open access to data and code}
    \item[] Question: Does the paper provide open access to the data and code, with sufficient instructions to faithfully reproduce the main experimental results, as described in supplemental material?
    \item[] Answer: \answerYes{} 
    \item[] Justification: We have attached the Github repository link including all code and data. The main page of the repository contains a detailed instruction to replicate all results in this paper. 

\item {\bf Experimental setting/details}
    \item[] Question: Does the paper specify all the training and test details (e.g., data splits, hyperparameters, how they were chosen, type of optimizer, etc.) necessary to understand the results?
    \item[] Answer: \answerYes{} 
    \item[] Justification: We clearly specifies the training and test settings used for the experiments. We provides details on data splits (e.g., 70/10/20 for train/validation/test), the optimizer (Adam), learning rate (0.005), number of training epochs (1000), model architecture (number of layers and hidden dimensions), and the procedure for hyperparameter tuning, including the regularization strength. Evaluation is conducted under multiple random seeds, and the criteria for model selection (based on validation performance) are explicitly stated. These details are sufficient to understand and interpret the reported results.

\item {\bf Experiment statistical significance}
    \item[] Question: Does the paper report error bars suitably and correctly defined or other appropriate information about the statistical significance of the experiments?
    \item[] Answer: \answerYes{} 
    \item[] Justification: For every experimental setting we reports \emph{mean ± standard‑deviation} of the evaluation metric over 10 random seeds (Table 2).  We explicitly state that each hyper‑parameter configuration is run with 10 different random initializations on a fixed node splits.

\item {\bf Experiments compute resources}
    \item[] Question: For each experiment, does the paper provide sufficient information on the computer resources (type of compute workers, memory, time of execution) needed to reproduce the experiments?
    \item[] Answer: \answerYes{} 
    \item[] Justification: 
    We have provided the CPU/GPU/memory information and the cloud computation platform to run the experiment in the Technique Appendices part. We have also provided the amount o compute in terms of running time to run an experiment setting and the total compute can be calculated by multiplying the running time under one setting by the total number of settings.  
    
\item {\bf Code of ethics}
    \item[] Question: Does the research conducted in the paper conform, in every respect, with the NeurIPS Code of Ethics \url{https://neurips.cc/public/EthicsGuidelines}?
    \item[] Answer: \answerYes{} 
    \item[] Justification: 

\item {\bf Broader impacts}
    \item[] Question: Does the paper discuss both potential positive societal impacts and negative societal impacts of the work performed?
    \item[] Answer: \answerYes{} 
    \item[] Justification: We have discuss the potential positive societal impacts in biology field in the discussion section. And we also include a potential negative societal impact in the appendix.
    
\item {\bf Safeguards}
    \item[] Question: Does the paper describe safeguards that have been put in place for responsible release of data or models that have a high risk for misuse (e.g., pretrained language models, image generators, or scraped datasets)?
    \item[] Answer: \answerNA{}{} 
    \item[] Justification:

\item {\bf Licenses for existing assets}
    \item[] Question: Are the creators or original owners of assets (e.g., code, data, models), used in the paper, properly credited and are the license and terms of use explicitly mentioned and properly respected?
    \item[] Answer: \answerYes{} 
    \item[] Justification:

\item {\bf New assets}
    \item[] Question: Are new assets introduced in the paper well documented and is the documentation provided alongside the assets?
    \item[] Answer: \answerNA{} 
    \item[] Justification: 
    
    Datasets: Cora, CiteSeer, and PubMed datasets are used and cited appropriately ([26]), which are publicly available benchmark citation networks commonly used in GNN research.

    Protein and metabolite data: we use protein–protein interaction data from the STRING database and metabolite–protein interaction data from KEGG, both of which are credited ([31], [35]).
    
    Protein embedding model: The ProtT5 model for protein sequence embedding is credited ([32]), and the source (UniProt) is mentioned for downloading sequences.
        
    Molecular fingerprints: Metabolite features are generated using RDKit, which is an open-source cheminformatics toolkit and is cited ([33]).

\item {\bf Crowdsourcing and research with human subjects}
    \item[] Question: For crowdsourcing experiments and research with human subjects, does the paper include the full text of instructions given to participants and screenshots, if applicable, as well as details about compensation (if any)? 
    \item[] Answer: \answerNA{} 
    \item[] Justification: \answerNA{}

\item {\bf Institutional review board (IRB) approvals or equivalent for research with human subjects}
    \item[] Question: Does the paper describe potential risks incurred by study participants, whether such risks were disclosed to the subjects, and whether Institutional Review Board (IRB) approvals (or an equivalent approval/review based on the requirements of your country or institution) were obtained?
    \item[] Answer: \answerNA{} 
    \item[] Justification: 

\item {\bf Declaration of LLM usage}
    \item[] Question: Does the paper describe the usage of LLMs if it is an important, original, or non-standard component of the core methods in this research? Note that if the LLM is used only for writing, editing, or formatting purposes and does not impact the core methodology, scientific rigorousness, or originality of the research, declaration is not required.
    \item[] Answer: \answerNA{} 
    \item[] Justification: \answerNA{}

\end{enumerate}

\end{document}